# Thermo-LIO: A Novel Multi-Sensor Integrated System for Structural Health Monitoring

Chao Yang[*1], Haoyuan Zheng[*1] and Yue Ma[#2]

**Xi'an Jiaotong Liverpool University**

**Abstract**—Traditional two-dimensional thermography, despite being non-invasive and useful for defect detection in the construction field, is limited in effectively assessing complex geometries, inaccessible areas, and subsurface defects.

This paper introduces Thermo-LIO, a novel multi-sensor system that can enhance Structural Health Monitoring (SHM) by fusing thermal imaging with high-resolution LiDAR. To achieve this, the study first develops a multimodal fusion method combining thermal imaging and LiDAR, enabling precise calibration and synchronization of multimodal data streams to create accurate representations of temperature distributions in buildings. Second, it integrates this fusion approach with LiDAR-Inertial Odometry (LIO), enabling full coverage of large-scale structures and allowing for detailed monitoring of temperature variations and defect detection across inspection cycles.

Experimental validations, including case studies on a bridge and a hall building, demonstrate that Thermo-LIO can detect detailed thermal anomalies and structural defects more accurately than traditional methods. The system enhances diagnostic precision, enables real-time processing, and expands inspection coverage, highlighting the crucial role of multimodal sensor integration in advancing SHM methodologies for large-scale civil infrastructure.

## 1. Introduction

Structural health monitoring (SHM) using thermography is a powerful non-destructive inspection method, particularly for critical infrastructure such as bridges [1,2], concrete walls [3], and historic buildings [4,5]. It effectively reveals hidden defects by capturing temperature variations indicative of underlying issues [9]. Infrared cameras detect temperature variations by converting infrared radiation emitted from the structural surface into electrical signals [1]. This process allows for the identification of damaged areas through temperature distribution patterns, which are governed by the Stefan-Boltzmann law [2, 3].

[1] These authors contributed equally to this work.
[1] C.Yang: 3368058743@qq.com  H.Zheng: Haoyuan.Zheng.xjtlu@gmail.com
[2] Corresponding author: Yue Ma02@xjtlu.edu.cn

Despite its widespread application in various civil structures inspection applications, thermography has several challenges that may severely impact its accuracy and effectiveness. [4]. Two-dimensional thermography, while effective for detecting surface temperature variations, has several notable limitations. Firstly, it requires a clear line of sight to the inspected area, which restricts its effectiveness with limited access or when inspecting structures with intricate designs or multiple layers. [5]. Secondly, it struggles to accurately interpret complex geometries. In such cases, thermal images may not fully reflect the true condition of these areas, leading to potentially distorted assessments. [6]. Additionally, since it only captures surface temperature differences, two-dimensional thermography cannot provide precise defect location information in three-dimensional space. Therefore, fusing 3D information into thermographic inspections, forming a new multimodality data format, is crucial for overcoming these limitations, enabling more accurate defect detection and providing a comprehensive assessment of the full extent of structural issues.

To address these critical gaps, this study introduces Thermo-LIO, a novel SHM system to integrate thermal imaging with high-resolution LiDAR technology to accurately identify and spatially localize thermal anomalies and structural defects within large-scale structures. Our contribution is threefold,

- The system generates calibrated three-dimensional thermographic models by precisely fusing temperature data with structural information.
- This capability enables detailed statistical structural analysis and provides new insights for intelligent monitoring of large-scale civil infrastructure.
- Developing and implementing hardware together with a complete software kit to conduct experiments in several typical scenes.

## 2. Related works

Current thermography inspections for outdoor SHM are mainly based on two-dimensional passive thermography, which uses the natural thermal radiation emitted by structures without the need for an external heat source [7-9]. This method is widely used because it does not require active thermal excitation and can cover large areas. Thermal cameras can be positioned on the ground for fixed-angle monitoring or mounted on mobile platforms for dynamic inspections [10], offering the flexibility to monitor structures from distances of 5-20 meters [18]. For example, Matsumoto et al. [11] used thermocouples to measure temperature variations on the bridge deck, accounting for factors such as solar radiation, atmospheric temperature, and monitoring periods. T. Omar et al. [12] used a mobile platform-mounted thermal imaging camera to inspect the bridge structure. By pre-processing infrared images and applying machine learning, they classified temperature data into different regions and created a new regional classification map. Kulkarni et al. [13] presented a framework that combines thermography with an EfficientNet to detect potholes in road pavements using infrared images captured by Unmanned Aerial Vehicles (UAVs). This approach addressed the limitations of traditional thermography, such as low resolution and signal-to-noise issues. However, these studies overlooked the structural integrity of the bridge deck, a 3D area whose temperature distribution may reflect the stress concentration evolution. The lack of detailed thermal information in the 3D geometry results in an incomplete assessment of the entire structure. Xiang Li et al. introduced a novel thermography method for predicting the 3D surface temperature field [14]. This approach, based on the restored pseudo-heat flux theory, simulates temperature differences between subsurface defects and intact areas. The method was validated through experiments on materials such as glass fiber reinforced plastic (GFRP), carbon fiber reinforced plastic (CFRP), and rubber, and it enables the assessment of defect characteristics such as depth, radius, and diameter-to-depth ratio, thus enhancing the design of nondestructive testing (NDT) using thermography. However, the method requires continuous heat excitation, and it focuses on small-sized samples. For outdoor and large-scale infrastructures, the structures receive varied environmental excitation, and the thermal flow is unpredictable.

With the rapid development of geometry reconstruction methods by image processing, thermographic data, when integrated with advanced technologies such as thermal camera arrays [15], is able to cover a wider range of structures and improve the inspection performance, offering a range of more comprehensive two-dimensional thermographic methods. Lagüela et al. [16] developed a thermographic model with depth information by integrating infrared and visible images, merging geometric data with thermal information to generate a detailed thermographic representation of structures. To maintain consistency during image registration and the construction of the thermographic mosaic, they utilized a uniform threshold across all images. However, the thermal view was fixed to a single view-angle, limiting its performance in building with more complex geometry. Daffara et al. [17] introduced an aerial thermography system designed for building analysis, utilizing UAVs fitted with dual visible-thermal sensors. Their approach involved the capture of synchronized thermal and visible images, which were then geometrically calibrated to ensure accurate alignment of thermal data onto structures generated using Structure from Motion (SFM) and Multi-View Stereo (MVS) techniques. However, these method relies on the off-line registration of image sequences, which requires expensive computational cost and cannot achieve real-time inspection, limiting their effectiveness in many in-situ monitoring applications.

Recent advancements in LiDAR have significantly expanded the potential applications of SHM on civil structures. [18]. By combining thermal data with 3D LiDAR, SHM becomes possible to accurately localize thermal anomalies within the physical space of civil structures. [19]. This integration allows for a more detailed understanding of structural issues, as it not only identifies anomalies but also places them in their precise spatial context. Biscarini et al. [20] Employed a combination of UAV photogrammetry, infrared thermography, and Ground Penetrating Radar (GPR) to assess the structural integrity of the Roman masonry bridge, Ponte Lucano. This approach enables precise identification of thermal anomalies associated with moisture and material degradation, in conjunction with subsurface features revealed by GPR with a detailed 3D model. Fang et al. [21] used multiple sensing technologies, such as 3D laser scanning, close-range photogrammetry, and infrared thermal imaging to produce detailed data of temple architecture. This methodology deployed both spatial and thermal data into a model, identifying structural deformations and potential areas of deterioration, but the data from multiple sensors was not fused. Therefore, the temperature distribution in the three-dimensional space cannot be presented. Hence, such conventional methods for large-scale SHM face significant limitations. Monitoring inaccessible areas is challenging, often resulting in incomplete coverage and the potential oversight of critical defects. Besides, the common practice of inspecting structures region-by-region fails to provide the holistic perspective necessary to fully assess defect extent and accurately evaluate overall structural deformation. Compounding these issues, the requirement for multiple sensor setups and varied viewing angles drastically reduces inspection efficiency, increasing time and cost burdens while compromising the effectiveness and reliability of thermographic SHM.

To overcome these challenges and achieve robust structural evaluation, a comprehensive thermographic model is essential. Such advanced thermal imaging is crucial for revealing the underlying mechanisms driving defect formation in civil infrastructure.

## 3. Methodologies

### 3.1 Multimodality fusion

The Thermo-LIO system integrates multiple sensing modalities through a structured development pipeline, as illustrated in Figure 1. It begins with radiometric calibration of the thermal camera to enable quantitative temperature measurement. The multi-modal data acquisition platform concurrently captures thermal images and 3D point clouds, with real-time data recording and processing handled by an embedded computer. After performing intrinsic and extrinsic sensor calibration along with temporal and spatial synchronization, the system applies distortion correction to the reconstructed 3D model. This integrated pipeline enables real-time visualization of temperature-mapped 3D models through a dedicated interface.

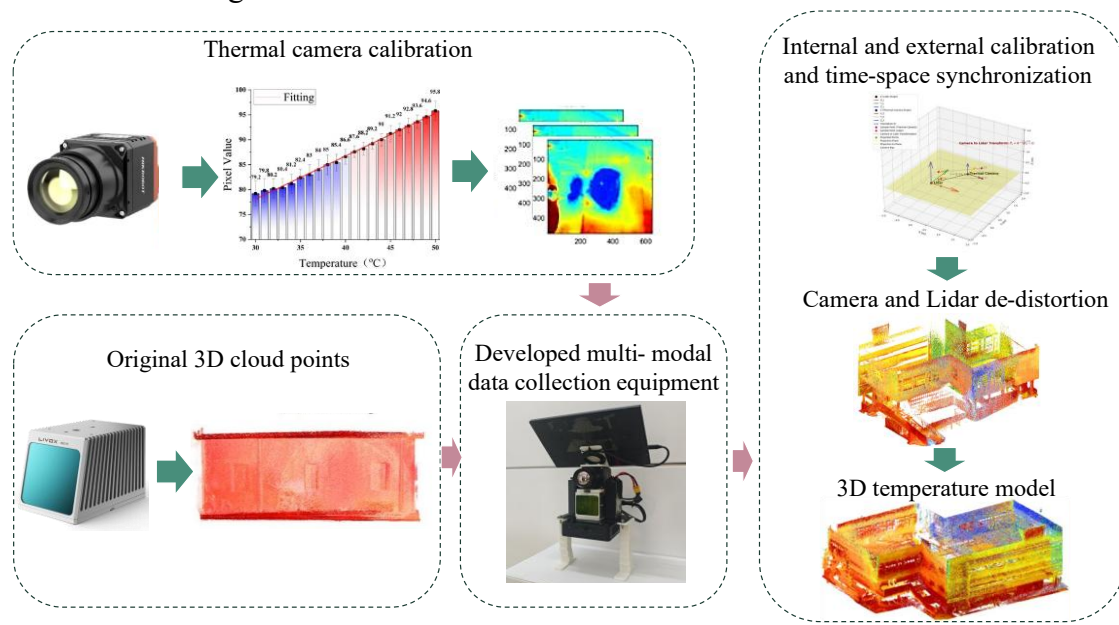

Figure 1. Framework for the development of the Thermal-Lio system

### 3.2 Calibration of Thermal Imaging
**(a) Temperature Model Calibration**

Radiometric calibration is essential to convert raw thermal pixels into accurate temperature values for seamless fusion with point cloud data. In this study, radiometric calibration is conducted to establish a quantitative relationship between the camera's digital output (gray values) and absolute temperature within a specified range of 30 to 50 °C. As shown in Figure 2, a high-emissivity (>0.99) extended-area blackbody reference source was positioned perpendicular to the thermal camera's optical axis at a fixed distance of approximately 1 meter to ensure uniform illumination and minimize parallax error. Thermal equilibrium was maintained at each calibration temperature, with stabilization periods exceeding 15 minutes per point to mitigate temporal drift. At each temperature plateau, multiple radiometric frames were captured using the camera's native software with fixed integration time and gain settings. For each set point, the mean gray value (in DN—Digital Number) was calculated from a central Region of Interest (ROI) covering at least 80% of the blackbody's emissive surface to maximize the signal-to-noise ratio. The precise absolute temperature of the blackbody surface at the time of image capture—verified to within ±0.1 °C using an integrated, traceable thermistor—was recorded as the ground truth reference.

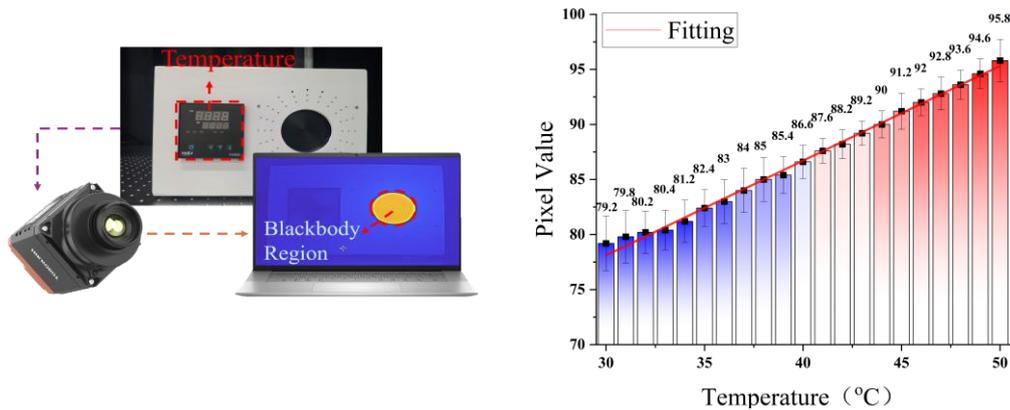

Figure 2. Thermal camera calibration experiment. (a) A calibrated thermal camera and the external black body source. (b). Temperature calibration error

Theoretical spectral radiance $L_\lambda$ $(T)$ was calculated for each reference temperature $T$ Using Planck's law, integrated over the camera's characterized spectral response band (8-14 μm). A linear least-squares regression model, $DN = K \cdot L(T) + B$, where $K$ represents the system's responsivity $(DN/W \cdot m^{-2} \cdot sr^{-1})$ and $B$ is an offset $(DN)$, was fitted to the paired data of average $DN$ values versus calculated radiance $L(T)$. Validation involved applying the derived calibration equation to intermediate temperatures (e.g., 32 °C, 38 °C, 48 °C) not used during fitting; measured temperatures showed deviations consistently below ±0.2 °C relative to the blackbody reference standard within the calibrated range. Ambient laboratory conditions (temperature: 22 ± 1°C, relative humidity: 50 ± 5%) were continuously monitored throughout the calibration procedure.

**(b) Thermal imaging calibration**
Infrared imaging can be described using the pinhole camera model, which defines the projection relationship between three-dimensional space points and their corresponding image pixels as follows:

$$s \begin{bmatrix} u \\ v \\ 1 \end{bmatrix} = K[R \quad t] \begin{bmatrix} X_W \\ Y_W \\ Z_W \\ 1 \end{bmatrix} \quad (1)$$

where $s$ is the scaling factor, $K$ is the intrinsic matrix, and $[R, t]$ Is the extrinsic matrix, respectively. Besides, for the nonlinear distortions, the radial distortions and tangential distortions are modeled as:

$$\begin{cases} x_{radial} = x(1+k_1 r^2 + k_2 r^4 + k_3 r^6) \quad y_{radial} = y(1+k_1 r^2 + k_2 r^4 + k_3 r^6) \\ x_{tangential} = x + [2p_1 xy + p_2(r^2 + 2x^2)] \quad y_{tangential} = y + [p_1(r^2 + 2y^2) + 2p_2 xy] \end{cases} \quad (2)$$

where $(x, y)$ are normalized image coordinates, $r^2 = x^2 + y^2$, $k_i$ is the radial distortion coefficient, and $P_i$ denotes the tangential distortion coefficients. These coefficients form the complete distortion vector. $[k_1, k_2, p_1, p_2, k_3]$.

We follow the calibration method for RGB camera calibration [22], which captures a checkerboard at different view angles to perform calibration. Given the corner points of a checkerboard in world coordinates $(X_w, Y_w, 0)$, the projection simplifies to:

$$s \begin{bmatrix} u \\ v \\ 1 \end{bmatrix} = K[r_1 \quad r_2 \quad t] \begin{bmatrix} X_w \\ Y_w \\ 1 \end{bmatrix} \quad (3)$$

Constraints from the rotation matrix columns $r_1$, $r_2$ yield equations that solve for the intrinsic matrix using linear least squares, followed by nonlinear optimization to refine the intrinsic parameters:

$$\min_{K, R_i, t_i} \sum_{i,j} \left\| m_{ij} - \hat{m}(K, R_i, t_i, M_j) \right\|^2 \quad (4)$$

Here, $m_{i,j}$ and $\hat{m}$ are observed and reprojected corner coordinates, respectively.

### 3.3 Calibration between Sensing Modalities

**(a) Coordinate transformation**

The coordinate systems of the multi-modal fusion are defined as: the LiDAR coordinate system $L$, Thermal camera coordinate system $C$, as shown in Figure 3. The extrinsic calibration between LiDAR and thermal imaging is denoted as $^{C}_{L}T$.

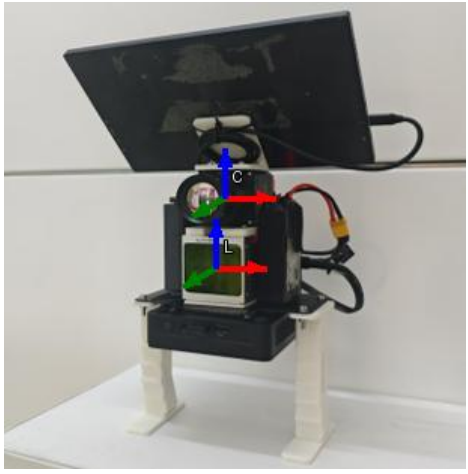

Figure 3. Transformation of thermal camera coordinates to LiDAR coordinates in 3D space.

Conventional LiDAR-RGB calibration methods project point clouds onto the image plane but suffer from occlusion-induced multi-value and null-value mappings. To overcome this, we directly extract edge features from LiDAR data and match them with infrared edges, eliminating the need for calibration boards and enabling use in outdoor scenes. Of the two edge types, depth-discontinuous and depth-continuous, only the latter is reliable, as depth-discontinuous edges are affected by laser beam divergence and mixed reflections at depth boundaries, which artificially enlarge foreground objects and introduce significant errors [30]. Infrared cameras present an additional challenge: they detect thermal gradients rather than visible-light intensity gradients. Therefore, to ensure robust calibration and avoid errors associated with depth-discontinuous edges, a modified Canny edge detection algorithm was used to capture the edges in infrared thermal imagery.

The extracted LiDAR edges must be matched with their corresponding edges in the infrared images. For each LiDAR edge, multiple points are sampled along with its geometric contour. Each sampled point, denoted as $LP_i \in \mathbb{R}^3$ in the LiDAR coordinate system, is projected into the camera frame using the current extrinsic $^{C}_{L}\overline{T}$ estimates $CP_i = {^{C}_{L}}\overline{T}(^{L}P_i) \in \mathbb{R}^3$. The transformed points $CP_i$ are subsequently projected onto the camera image plane, generating theoretical projection coordinates $Cp_i = \pi(^{C}P_i)$, where $\pi(P)$ denotes the perspective projection operation. However, due to inherent lens distortion in physical cameras, the actual observed position of the projected point on the image plane p_i=(u_i,v_i) is modeled as $p_i = f({}^Cp_i)$, where $f(p)$ is the camera distortion model. During the k-d tree-based nearest neighbor $p_i$ search, let $Q_i = q_i^j; j = 1, \cdots, \kappa$ denote the retrieved k-nearest neighbors, which are incorporated into the following equation for computation:

$$q_i = \frac{1}{\kappa}\sum_{j=1}^{\kappa} q_i^j; \quad S_i = \sum_{j=1}^{\kappa}(q_i^j - q_i)(q_i^j - q_i)^T \tag{5}$$

Then, the line formed. $Q_i$ is parameterized by a point $q_i$ and a normal vector $n_i$, where $S_i$ is the eigenvector corresponding to the smallest eigenvalue $n_i$.

For the infrared camera, thermal gradient variations in the image must also be considered, as they may not fully coincide with traditional geometric edges. Therefore, in addition to projecting the sampled points from LiDAR edges $LP_i$ onto the image plane, the edge direction vector was projected to verify its orthogonality to the thermal gradient vector $n_i$. When two non-parallel lines in the image plane exhibit spatial proximity, incorrect matches can be effectively eliminated.

**(b) Noise Analysis of LiDAR Measurement**

The extracted LiDAR edge points $LP_i$ and their corresponding edge features $(n_i, q_i)$ in the image are affected by measurement noise. Therefore, a comprehensive noise model must be established. First, let $Iw_i \in \mathcal{N}(0, {}^I\Sigma_i)$ represent the noise associated with the $q_i$, with its covariance matrix $I\Sigma_i = \sigma_I^2 I_{2\times 2}$, where $\sigma_I = 1.5$ denotes the pixel noise caused by pixel discretization. For a LiDAR point $LP_i$, let $\omega_i \in \mathbb{S}^2$ be the measured azimuthal direction, and $\delta\omega_i \sim \mathcal{N}(0_{2\times 1}, \Sigma_{\omega_i})$ be the measurement noise on the tangential plane $\omega_i$. Using the operator of $\mathbb{S}^2$, the relationship between the true azimuthal direction $\omega_i^{gt}$ and its measured value $\omega_i$ is expressed as:

$$\omega_i^{gt} = \omega_i(_{\mathbb{S}^2}\delta\omega_i \triangleq e^{[N(\omega_i)\delta\omega_i \times]}\omega_i \tag{6}$$

where $N(\omega_i) = [N_1; N_2] \in \mathbb{R}^{3\times 2}$ is the orthonormal basis of the tangential plane at $\omega_i$, and $[\cdot \times]$ represents the cross-product matrix. The (operator on $\mathbb{S}^2$ essentially rotates the unit vector $\omega_i$ around an axis on the tangential plane $\omega_i$ by $\delta\omega_i$, ensuring the result remains a unit vector (i.e., stays on the tangential plane $\mathbb{S}^2$).

The targetless-based method calibrates the LiDAR-thermal camera extrinsic parameters using scenes rich in geometric features. Careful scene selection is crucial: purely cylindrical objects like pillars are unsuitable as they hinder plane fitting for edge extraction. For effective calibration, the extracted edges must distribute relatively evenly across the Field of View (FoV) to mitigate noise and exhibit multi-directionality (avoiding unidirectional patterns like purely vertical edges). An initial extrinsic matrix, derived from the CAD model, is optimized through edge matching. Figure 4 illustrates this method, showing an original RGB image, extracted features, LiDAR depth projected onto the thermal image, and the colorized point cloud.

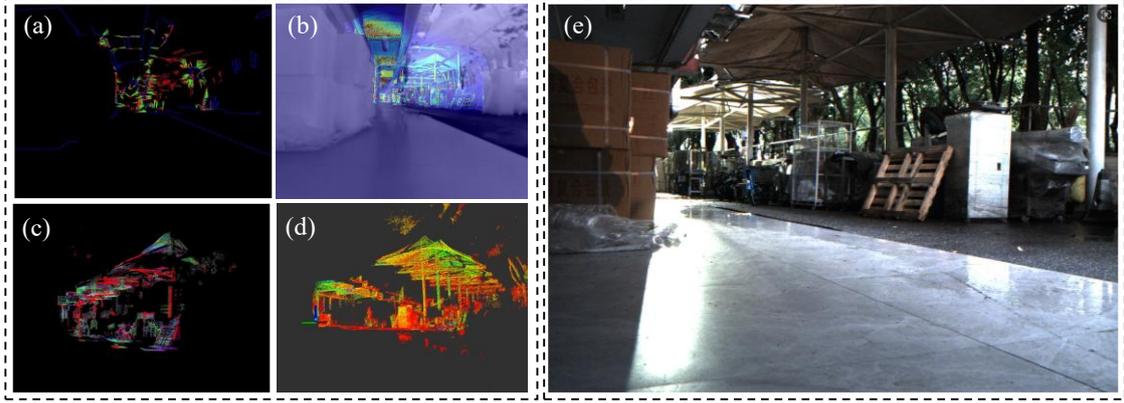

Figure 4. Calibration example

Similarly, let $d_i$ be the measured depth and $\delta d_i \sim \mathcal{N}(0, \Sigma_{d_i})$ be the ranging error. The true depth $d_i^{gt} = d_i + \delta d_i$, Combining the above equations, the relationship between the true point position $LP_i^{gt}$ and its measured value $LP_i$ is:

$$LP_i^{gt} = d_i^{gt}\omega_i^{gt} = (d_i + \delta d_i)(\omega_i \boxplus_{\mathbb{S}^2} \delta\omega_i) \approx d_i\omega_i + \omega_i\delta d_i - d_i[\omega_i \times]N(\omega_i)\delta\omega_i = \underbrace{d_i\omega_i}_{LP_i} + \underbrace{\omega_i\delta d_i - d_i[\omega_i \times]N(\omega_i)\delta\omega_i}_{Lw_i} \quad (7)$$

Therefore, the noise of the LiDAR point $LP_i$ is:

$$L_{w_i} = [\boldsymbol{\omega}_i - d_i[\boldsymbol{\omega}_i \times]N(\boldsymbol{\omega}_i)]\begin{bmatrix}\delta d_i \\ \delta\boldsymbol{\omega}_i\end{bmatrix} \sim \mathcal{N}(0, L_{\Sigma_i}) \quad (8)$$

Where:

$$L_{\Sigma_i} = A_i \begin{bmatrix}\Sigma_{d_i} & 0_{1\times2} \\ 0_{2\times1} & \Sigma_{\boldsymbol{\omega}_i}\end{bmatrix} A_i^T, \quad A_i = [\boldsymbol{\omega}_i - d_i[\boldsymbol{\omega}_i \times]N(\boldsymbol{\omega}_i)] \quad (9)$$

This noise model will be employed to enable consistent extrinsic calibration in subsequent procedures.

**(c) Calibration Formulation and Optimization**

Let $LP_i$ denote an edge point extracted from the LiDAR point cloud, and let its corresponding edge in the image be represented by a normal vector $n_i \in \mathbb{S}^1$ and a point $q_i \in \mathbb{R}^2$ on the edge. Considering the noise in $LP_i$ and projecting it onto the image using the true extrinsic, the projected point should ideally lie precisely on the edge $(n_i, q_i)$ extracted from the image. This constraint is expressed as:

$$0 = n_i^T (f(\pi(^C_L T(^L P_i + ^L w_i))) - (q_i + ^I w_i)) \tag{10}$$

Where the $Lw_i \in \mathcal{N}(0, ^L \Sigma_i)$ 和 $Iw_i \in \mathcal{N}(0, ^I \Sigma_i)$. This nonlinear system is solved iteratively via perturbation-based optimization on the SE(3) manifold. Let $^C_L \bar{T}$ denote the current extrinsic parameter estimate:

$$^C_L T = ^C_L \bar{T}(_{SE(3)} \delta T \triangleq \mathrm{Exp}(\delta T) \cdot ^C_L \bar{T} \tag{11}$$

Where:

$$\delta \mathcal{T} = \begin{bmatrix} \delta \boldsymbol{\theta} \\ \delta \boldsymbol{t} \end{bmatrix} \in \mathbb{R}^6; \quad \mathrm{Exp}(\delta \mathcal{T}) = \begin{bmatrix} e^{[\delta \boldsymbol{\theta} \times]} & \delta \boldsymbol{t} \\ \mathbf{0} & 1 \end{bmatrix} \in SE(3) \tag{12}$$

Substituting this into the equations and using a first-order approximation, we obtain $0 \approx r_i + J_{Ti} \delta T + J_{wi} w_i$. Then the optimal solution is derived via MLE as:

$$\delta T^* = -(J_T^T (J_w \Sigma J_w^T)^{-1} J_T)^{-1} J_T^T (J_w \Sigma J_w^T)^{-1} r \tag{13}$$

The extrinsic are iteratively updated until convergence by $^C_L \bar{T} \leftarrow ^C_L \bar{T}(_{SE(3)} \delta T^*$. Figure 5 demonstrates the comparison results between our method and several baseline methods regarding LiDAR-thermal rotation errors and LiDAR-thermal translation errors. Compared to other baseline methods [23]. The proposed approach exhibits lower errors across all three rotational dimensionsand all three translational dimensions (X, Y, Z), validating its high accuracy and robustness.

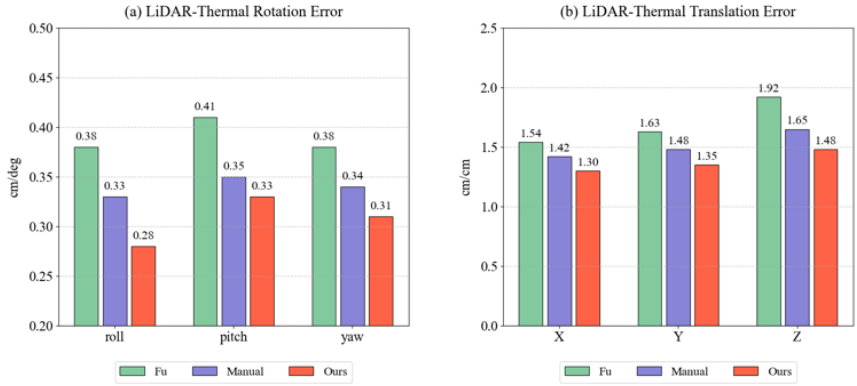

Figure 5. Comparison of calibration error

## 3.4 Data-fusion with LIO

The system employs an Iterative Extended Kalman Filter (IEKF) method, PointLIO [24] to achieve tight coupling between the lidar and IMU. As shown in Figure 6 (a), the hardware of the thermal-Lio system was developed and depicted, where it consists of an industrial long-wave infrared camera (MV-CI003-GL-N6) and a laser radar (Livox Mid-70). The LiDAR and thermal camera are installed together with a fixed configuration using 3D printing components. A microcomputer was employed to real-time analyze the collected 3D point and thermal image data and to generate the 3D thermal model, which was then demonstrated on the integrated screen. Unlike loosely coupled methods, this model directly fuses the lidar observation model with the IMU measurements during the state estimation, avoiding the accumulation of estimation errors. The main modules of the system include State estimation and Thermal Imaging Image Fusion. As shown in Figures 6 (b) and 6 (c), after the calibration, the handhold Thermal-LIO system was used to scan an outdoor structure with multiple views, and the point cloud data and thermal image data were collected at the same time. Following that, a 3D thermal model with Morphological and temperature characteristics can be generated, and the path of scanning is also recorded and demonstrated on the screen.

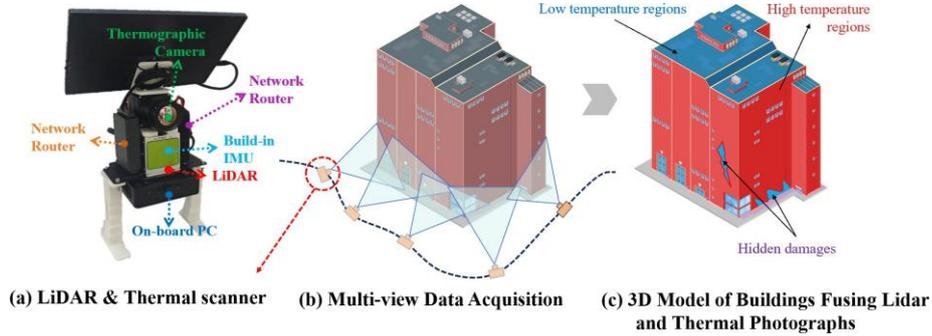

(a) LiDAR & Thermal scanner  (b) Multi-view Data Acquisition  (c) 3D Model of Buildings Fusing Lidar and Thermal Photographs

Figure 6. 3D temperature model data acquisition procedure with the Thermo-LIO system

The forward propagation algorithm is based on the IMU observation model. [25], and the system implements discrete state prediction using the following equation:

$$\hat{x}_{i+1} = \hat{x}_i \boxplus (\Delta t f(\hat{x}_i, u_{i,0}); \hat{x}_0 = \bar{x}_{k-1} \tag{14}$$

Where $f(\cdot)$ is a nonlinear motion model, and $\boxplus$ is a manifold addition operation. Covariance propagation uses an error state dynamics model:

$$\hat{P}_{i+1} = F_x \hat{P}_i F_x^T + F_w Q F_w^T; \hat{P}_0 = \overline{P_{K-1}} \tag{15}$$

Among them, $F_x$ and $F_w$ are the Jacobian matrices of the error state and process noise, respectively, which are efficiently calculated using automatic differentiation technology. This module achieves state prediction with O(1) time complexity.

Through the previous state estimation module, feature points in the lidar coordinate system are mapped to the global coordinate system using a multi-level coordinate transformation equation, as shown below:

$$^G\bar{p}_{fj} = {^G\bar{T}_{Ik}} \cdot {^I T_L} \cdot {^{L_k} p_{fj}}; j = 1,...,m \tag{16}$$

Among them, $^{L_k}\mathbf{p}_{f_j}$ denotes the j$_{th}$ feature point, the 3D coordinate vector of the j$_{th}$ feature point in the LiDAR coordinate system (L); $^I\mathbf{T}_L$ is the rigid transformation matrix from the laser radar to the IMU coordinate; $^G\bar{\mathbf{T}}_{I_k}$ is the pose transformation matrix from the IMU coordinate to the global coordinate (G). To generate a calibrated three-dimensional thermal imaging model, we fuse the thermal imaging images with a single frame of LiDAR scan by using intrinsic parameters from Section 3.2.

### 3.5 Implementation Details

Table 1. Technical parameters of thermal camera and LiDAR

| Industrial Long-wave IR Camera | Specification |
|---|---|
| IR resolution | 640 × 512 |
| Field of view (FOV) | 88.5° × 73.2° |
| Full image frequency | 50Hz |
| Focal length | 6.3 mm |
| Detector type | Vanadium oxide uncooled detector |
| Thermal sensitivity | <35 mk (F1.0, 25°C) |
| Laser Radar | Specification |
| Optical maser wavelength | 905nm |
| Field of view (FOV) | Non-repeat scan:70.4°x77.2° |
| | Repeat scan:70.4°x4.5° |
| Random error in ranging | 10(@ 20 m) <2 cm |
| Range (@0 klx) | 190m @ 10% reflectance |
| | 260m @ 20% reflectance |
| | 450m @ 80% reflectance |
| | 190m @ 10% reflectance |
| Range (@100 klx) | 230m @ 20% reflectance |
| | 320m @ 80% reflectance |

Some of the hardware parameters are presented in Table 1. An additive-manufactured handle structure was attached at the bottom of the system to allow the operator to scan outdoor structures. The intrinsic parameters of the thermal camera are evaluated before LiDAR-thermal camera calibration. During the LiDAR-thermal camera calibration, the system requires a data acquisition time of 15 seconds for SSL to obtain sufficiently dense points for feature extraction. The system typically requires less than 2 seconds to acquire sufficient points during visual sensing. Extrinsic calibration plays an essential role to accurately matching the depth and temperature stream from the LiDAR camera sensor.

## 4 Experiment and discussion

### 4.1. Experiment setup

Two scenes were selected for inspection. As shown in Figure 7, the first structure was the middle part of an overpass area, where the whole bridge structure was located over a wide car road. The overpass was used for pedestrians and bicycles to cross the road. The whole bridge contains concrete on the bottom and metal armrests. After long-term use and the action of rain, the bottom of the bridge has surface corrosion. At the same time, due to the sun exposure and the shadows created by the surrounding vegetation, the heat exchange between the building surface and the environment is uneven, resulting in the inability of normal infrared thermal imaging to analyze the temperature response of such a large outdoor structure. The second outdoor structure was a hall building located on the campus of a university. There are multiple kinds of defects over the whole building, such as concrete cracks and sludge resulting from the air conditioning leakage. In addition, different parts of the building received solar radiation from varied angles due to the motion of the sun. Therefore, a traditional single frame of thermal image from a fixed monitoring angle is not able to analyze the thermal evolution of the building over a day.

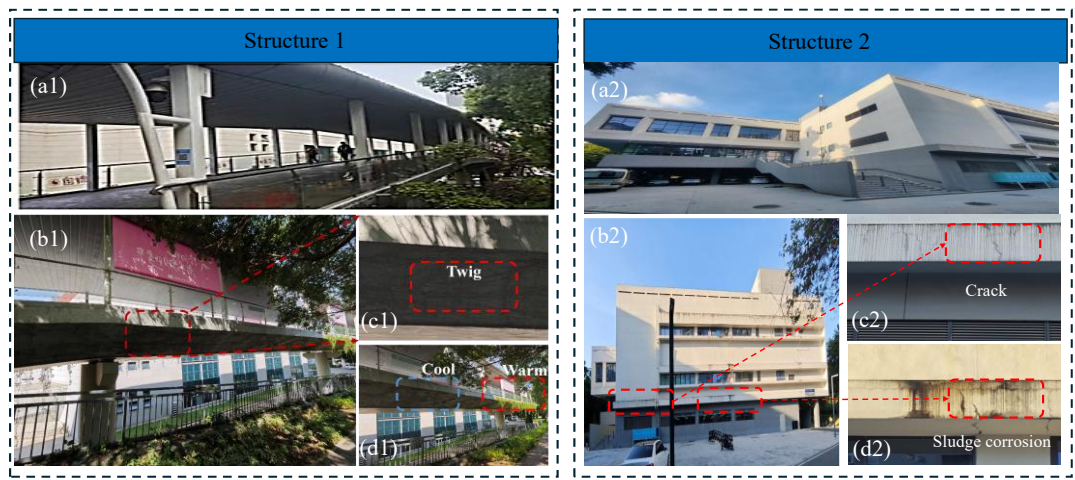

Figure. 7. Two inspected outdoor civil structures. (a) Overpass areas and local enlarged details; (b) building areas and local enlarged details

Table 2 shows the inspection parameters for the two outdoor structures. The inspection for the bridge part and the outer wall of the teaching building were conducted at 15 : 44 p.m. and 16 : 06 p.m on the same day in cloudy weather conditions respectively, and the highest temperature of the day was 33 °C and the lowest temperature was 26 °C. Using handheld devices, one side of the bridge and the external wall of dining hall building were scanned and modeled globally. The main materials of the two structures are concrete, cement and ceramic tiles. The emissivity of building materials at room temperature is close to 1, so the influence of ambient heat can be ignored. As shown in Figure 7(a), the left half of the bridge is shaded by trees to prevent all-day solar radiation, and the right half is exposed to sunlight. The two sections show obvious temperature differences, resulting in an increase in vegetation in the right half. The enlarged area shows that the concrete surface under the bridge is covered with vegetation, and small defects are easy to ignore. The outer wall of the teaching building is exposed to solar radiation for a long time, and small cracks appear on the wall, which is prone to the problem of porcelain falling off. There are also cracks and water seepage in some areas of the building. Due to the wet environment caused by water seepage, black sludge is attached to the wall. Compared with other areas, these areas with cracks and water seepage problems have suffered more serious degradation.

Table 2. Inspection parameters

|  | Structure 1 | Structure 2 |
| --- | --- | --- |
| Structure type | Overpass | Hall building |
| Location | Car road | University campus |
| Inspection time | 9:00-22:10 | 9:00-22:10 |
| Inspected region | Front | Whole building |
| Weather condition | Cloudy | Sunny |
| Local temperature | 26-33 °C | 26-31 °C |

### 4.2 Inspection results for the overpass

Figure 8 presents the 3D spatial temperature models of the overpass structure generated by the Thermo-LIO system, captured at four distinct times during a diurnal cycle (09:00, 12:45, 16:01, and 22:10). The results demonstrate the system's capability to capture the dynamic thermal evolution of a complex outdoor structure in real-time 3D. A striking spatial temperature variation is evident, with the left half of the bridge, shaded by trees, exhibiting consistently lower temperatures across all time points compared to the sun-exposed right half. This pronounced thermal contrast, resulting directly from uneven solar radiation due to the surrounding vegetation, highlights a significant environmental influence on the structure's thermal behavior that traditional fixed-view 2D thermography would struggle to contextualize spatially.

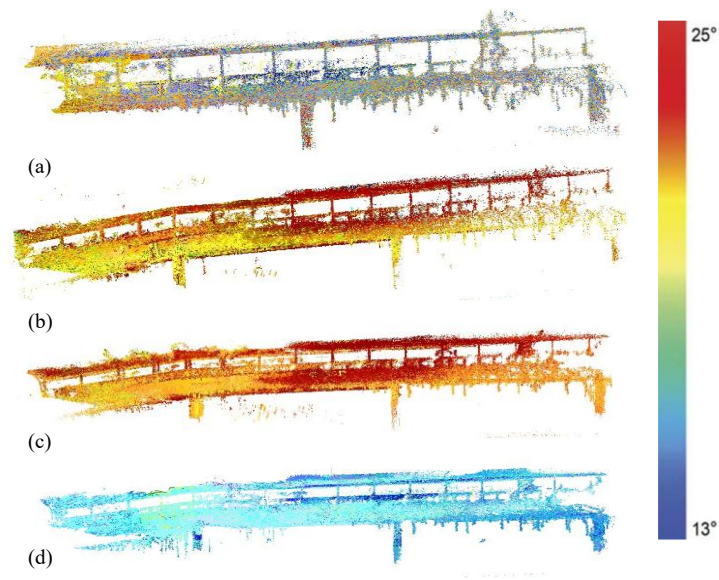

Figure 8. 3D spatial temperature model for the overpasses across diurnal cycles. (a) 09：00 (b). 12：45. (c). 16:01. (d). 22:10.

Furthermore, the sequence clearly illustrates the diurnal thermal loading cycle: temperatures rise from the morning (09:00) towards peak heating around midday and early afternoon (12:45, 16:01), followed by a significant cooling phase into the evening (22:10). During peak solar radiation at 12:45, significant temperature gradients were observed, particularly in areas directly exposed to sunlight compared to shaded regions. At 22:10, the thermal differences notably diminished but retained residual heat indicative of structural and material responses. Critically, the integration of thermal data onto the precise LiDAR-derived 3D geometric model allows for the accurate spatial localization of these thermal patterns relative to the bridge's physical features. This capability is essential for correlating observed thermal anomalies, such as potential subsurface defects hinted at by unusual temperature gradients in specific areas (like the vegetation-covered concrete underside shown in the enlarged detail), with their exact structural location, overcoming a fundamental limitation of conventional 2D thermography for assessing complex geometries and inaccessible areas like the bridge underside. The figure thus provides compelling evidence of Thermo-LIO's effectiveness in enabling detailed, spatially contextualized monitoring of thermal variations critical for structural health assessment over time.

**4.3 Inspection results for the building**

Apart from the monitoring results from a fixed orientation for the overpass, Figure 9 presents the 3D spatial temperature models of the university building generated by the Thermo-LIO system across the same diurnal cycle (09:19, 12:45, 16:01, 22:10) as the overpass in Figure 8. The results reveal complex spatial and temporal thermal patterns inherent to a large building facade under varying solar exposure. The models delineate significant temperature variations across different structural elements and materials, such as the distinct thermal signature of the brown glass door compared to the surrounding concrete walls. Crucially, the integrated 3D thermography enables the precise spatial localization of thermal anomalies indicative of structural defects, including areas affected by cracks, water seepage, and the resulting black sludge deposits mentioned in the case study. These degraded areas, subjected to prolonged moisture exposure, consistently exhibit lower temperatures compared to drier sections, a pattern visualized and mapped onto the building's geometry. The sequence further demonstrates the dynamic thermal response of the structure to solar loading: temperatures rise significantly by midday (12:45), with the building's thermal mass causing a lag in peak temperature observable in the early afternoon model (16:01), followed by substantial cooling by nightfall (22:10).

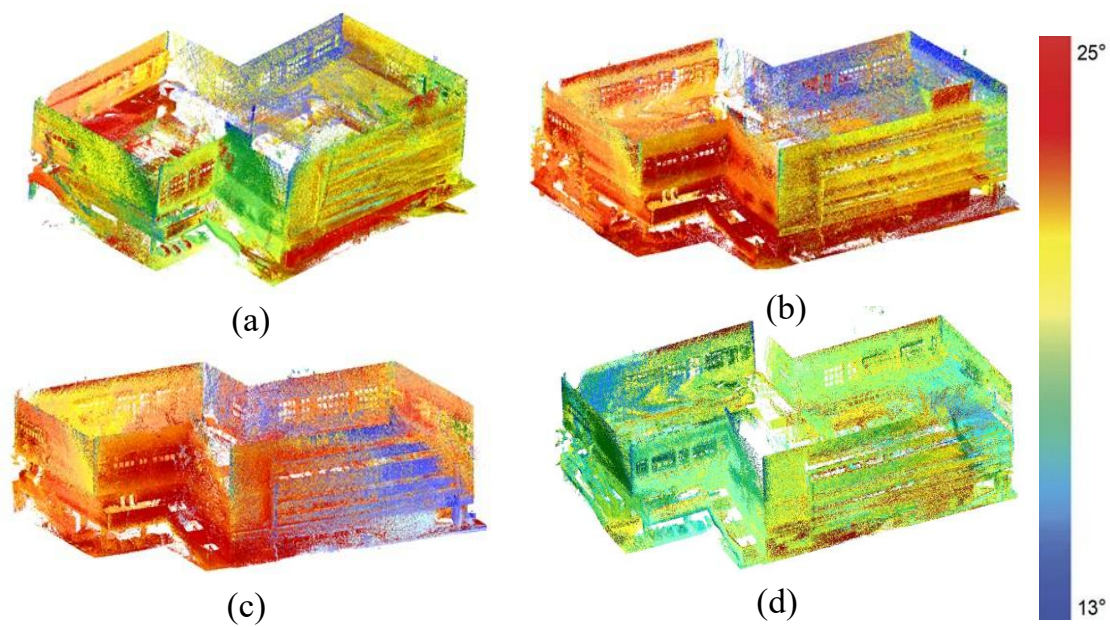

Figure 9. 3D spatial temperature model for the building across diurnal cycles. (a) 09：19 (b). 12:45. (c). 16:01. (d). 22:10.

The 3D visualization resolves the challenge posed by the sun's changing angle throughout the day, which creates complex, time-varying shadows and heating patterns that would confound single-viewpoint 2D thermography. The Thermo-LIO system successfully captures this evolution, allowing for correlation between specific thermal signatures (like persistent cool spots indicating moisture or delamination) and their exact 3D location on the building, even on surfaces not directly facing the initial scan position. This capability is vital for comprehensively assessing defect severity and planning targeted maintenance interventions.

**4.4 Monitoring results from varied orientations**

The capability of multi-perspective buildings was tested in this study as well. The visualizations collected from the Thermo-LIO system by three distinct orientations (Orientation 1, 2, and 3) in Figure 10 critically overcome the single-view limitation inherent in traditional thermography, revealing how thermal signatures and structural details vary significantly depending on the observation angle. The consistent temperature color mapping across all orientations allows for direct comparison, highlighting areas where thermal anomalies persist regardless of viewpoint, such as the persistently cooler regions associated with water seepage and sludge deposits mentioned in the case study, thereby confirming their significance as potential defect indicators rather than artifacts of a specific viewing geometry.

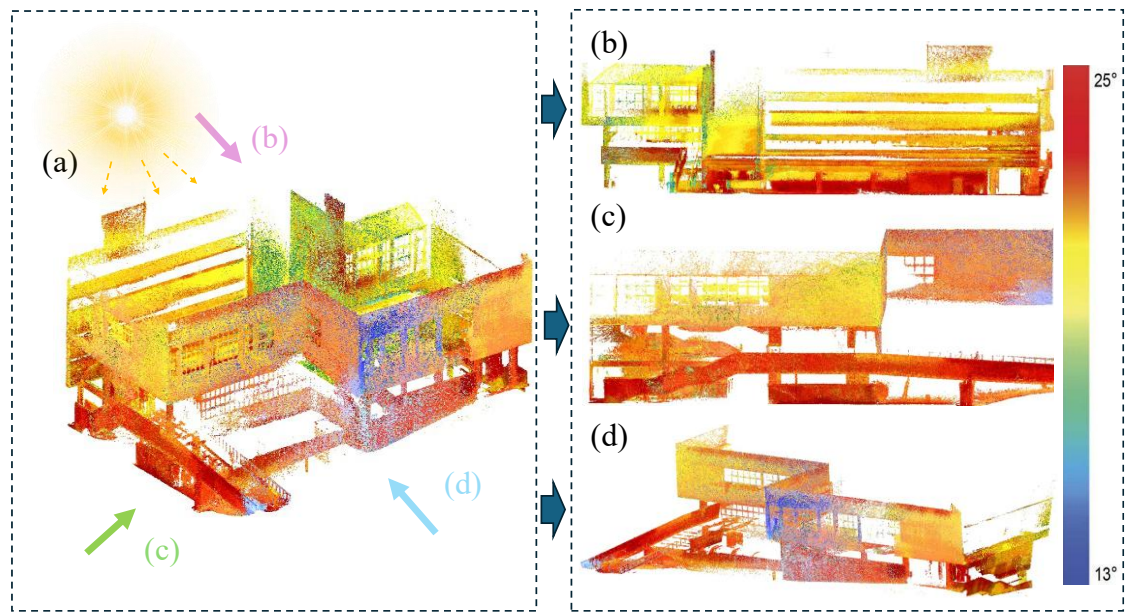

Figure 10. Different temperature models from varied orientations. (a) monitoring the orientations of the building. (b) temperature model from orientation 1 . (c) temperature model from orientation 2. (d) temperature model from orientation 3.

The system successfully integrates thermal data onto the complex 3D geometry captured by LiDAR, resulting in models that exhibit textural and thermal details closely resembling actual structural surfaces, as noted in the comparison to RGB imagery. This multi-angle reconstruction is particularly valuable for inspecting occluded or complex structural features, which might be partially hidden or distorted in a single thermal image. By synthesizing information from multiple scans, Thermo-LIO generates a comprehensive thermal representation of the entire structure, enabling the correlation of thermal patterns observed from one angle with their spatial context and structural relationships visible from another. This capability is essential for accurately diagnosing defects that manifest differently depending on perspective and for ensuring complete coverage in large-scale structural health assessments.

### 4.5 Segmented components within the global cloud points model

Figure 11 showcases the structural segmentation capability inherent in the Thermo-LIO system's 3D thermographic models, explicitly demonstrating how distinct architectural components of the building can be identified and isolated within the fused thermal-spatial dataset. The figure segments the comprehensive model Figure 11 (a) into key structural elements: the stair structure Figure 11 (b), the ramp driveway Figure 11 (c), and representative windows Fig 11 (d). This segmentation is not merely geometric; it critically incorporates the distinct thermal signatures associated with each component type. For instance, the stair structure, likely composed of concrete or metal, exhibits a characteristic thermal mass and potential heat transfer patterns influenced by its usage and exposure, while the ramp driveway shows a continuous surface with its thermal gradient profile, possibly affected by material composition (e.g., asphalt vs. concrete) and solar absorption. The windows are particularly illustrative, revealing the thermal behavior of glass, typically showing different emissivity, reflection, and insulation properties compared to opaque building materials, which manifests as unique temperature patterns around the frames and glazing.

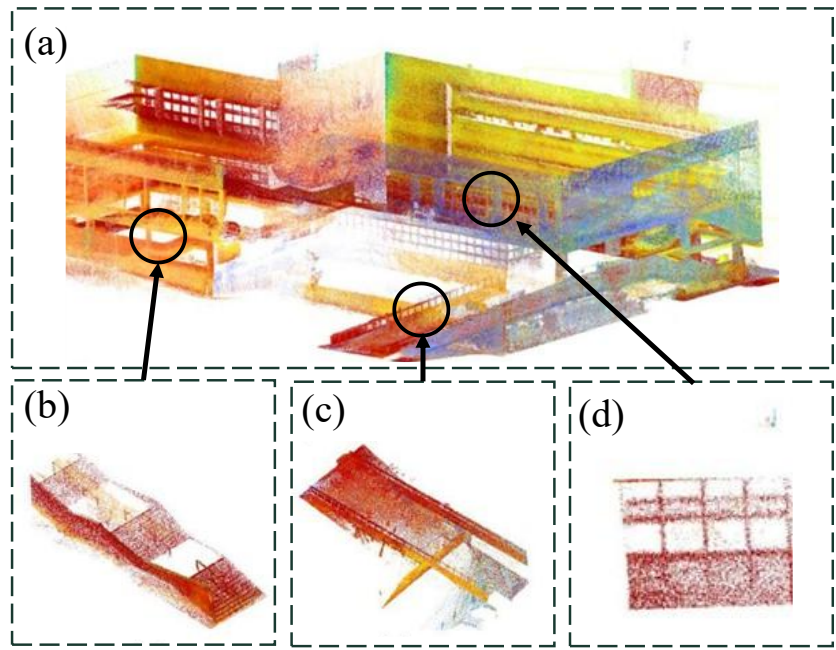

Figure 11. Segmented structure parts from the whole 3D temperature model of the building. (a) The whole 3D temperature model of the building. (b) Stair. (c) Ramp driveway. (c) Window.

The ability to resolve and isolate these specific structural elements within the global 3D thermographic model is paramount for targeted structural health assessment. It allows engineers to focus analysis on the thermal performance and potential defect indicators (like unusual heat loss around windows suggesting poor seals, or unexpected cold spots on stairs indicating potential subsurface moisture or delamination) specific to each component's function, material, and typical failure modes. This component-level resolution, enabled by the precise fusion of LiDAR geometry and thermal data, moves beyond holistic building assessment to provide actionable insights for component-specific maintenance planning and degradation analysis, significantly enhancing the diagnostic precision offered by the Thermo-LIO system. The developed 3D thermal model reconstruction method is compared with other state-of-the-art methods in Table 3.

Table 3. Comparison of the 3D temperature reconstruction methods

| Index | Our Method | Shin & Kim[24] | Chen et al. [26] | De Pazzi et al.[27] |
|---|---|---|---|---|
| Whether to build a 3D map in real time | ✓ | Sparse depth + Heat map | LiDAR point cloud+ Infrared edge map | Building heatmaps offline |
| Quantitatively analyze the heat map | Temperature inversion through a linear calibration model | Thermal image pixel values | Edge tracking | Thermal imager built-in software converts pixel intensity into apparent temperature values |
| Defect detection capability | Crack/water stain/peeling/ biological adhesion detection | Composition and positioning only | Composition and positioning only | -- |

**4.6 Enhanced cloud density at a fixed view**

To enhance the monitoring quality in a fixed angle and get an enhanced 3D spatial temperature model, the Thermal-Lio system is fixed in front of the monitoring area. Due to the static monitoring being conducted at a different time, the color of the model is slightly different from that of dynamic monitoring. Figure 12 demonstrates the superior resolution and thermal fidelity achievable with the Thermo-LIO system under fixed-orientation monitoring, contrasting with the previously discussed dynamic scans. By stabilizing the device in front of targeted structural regions (Regions 1 and 2), the system generates significantly denser point clouds Figure 12 (c) and (d) compared to handheld operation. This increased point density directly translates to enhanced geometric resolution, capturing finer structural details like subtle surface texture variations, minor cracks, or material interfaces that were potentially obscured or averaged out during dynamic acquisition.

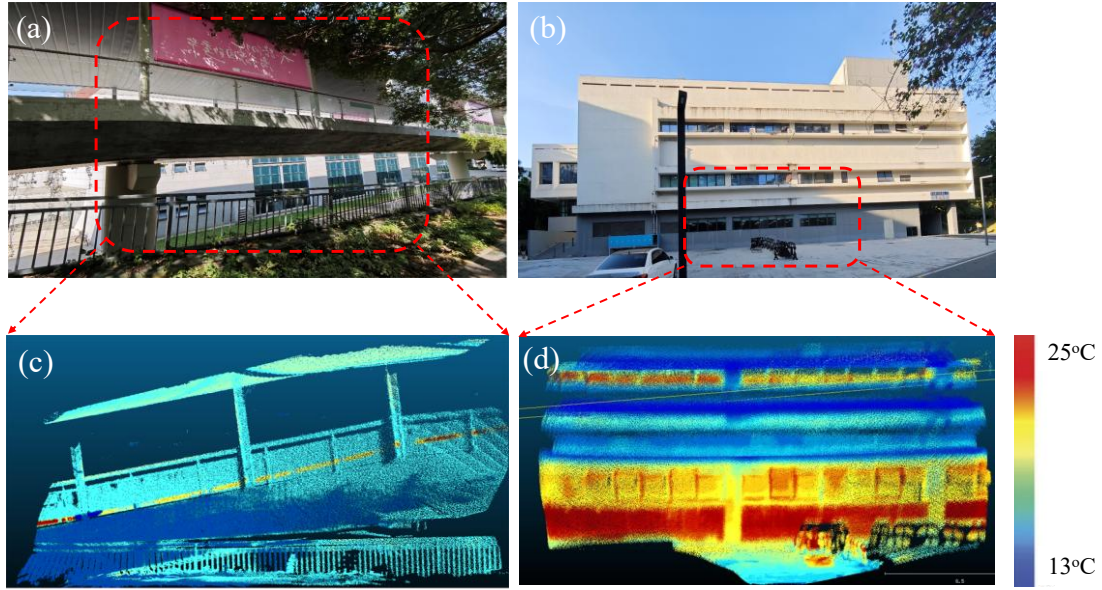

Figure 12. High-density point cloud temperature model derived from fixed monitoring orientation. (a) Monitored region from structure 1. (b) Monitored region from structure2. (c) High-density point cloud temperature model from the monitored region on (a). (d) High-density point cloud temperature model from the monitored region on (b).

The fixed position eliminates motion-induced artifacts and minimizes temporal inconsistencies inherent in moving scans, resulting in a smoother and more representative temperature transition across surfaces. The models exhibit reduced noise levels, particularly mitigating the disruptive infrared sensor noise triggered by automatic low-light illumination (a noted limitation during dynamic dusk/night scans). This stability allows the thermal camera to consistently sample the same area over time without perspective shifts, yielding a more accurate and continuous representation of the true thermal gradients across the structure. The resultant high-definition thermographic models provide the necessary precision for quantifying finer-scale thermal anomalies – such as pinpointing the exact boundaries of delamination, assessing heat flow around small cracks, or identifying localized moisture ingress with greater confidence – that are critical for early-stage defect diagnosis and detailed structural assessment. This fixed-view capability thus complements the system's mobile scanning, offering a high-resolution mode for targeted inspection of critical areas where maximum detail on both morphology and thermal response is paramount.

While advancing structural health monitoring through 3D thermographic fusion, the Thermo-LIO system still exhibits several limitations that warrant refinement. A primary drawback is its resolution constraints emerge during handheld operation: distant or fine-scale defects (e.g., microcracks under 1 mm) become undetectable as thermal camera resolution degrades beyond 10–15 meters, while LiDAR range limitations impede tall-structure coverage. Environmental adaptability also poses challenges. The dynamic nighttime scans trigger automatic infrared illuminators, injecting noise that disrupts thermal continuity. Computational demands for real-time 3D model generation further confine the system to embedded processing units (NUC), limiting deployment flexibility.

To address these gaps, future iterations will deploy higher-resolution thermal imagers and long-range LiDAR—which would improve defect detectability at scale. Drone-based deployment can also be considered for overcoming accessibility barriers, enabling inspections of elevated or confined areas (e.g., bridge undersides). In addition, Algorithmic refinements can suppress low-light noise artifacts and optimize edge computing pipelines to reduce latency. Transforming raw data into actionable structural insights. These advancements would expand Thermo-LIO's robustness, accuracy, and applicability across diverse infrastructure monitoring scenarios.

## 5  Conclusions

Given the limitations of conventional two-dimensional thermography in addressing complex geometries and subsurface defects, this paper proposes Thermo-LIO, a novel approach integrating thermal imaging with high-resolution LiDAR for advanced structural health monitoring. The primary aim was to develop a robust real-time inspection system capable of accurately identifying and localizing thermal anomalies within large-scale infrastructures. This was achieved through meticulous radiometric calibration, spatial synchronization of thermal and LiDAR sensors, and the creation of precise three-dimensional thermographic models. The key contributions of this research are summarized as below:

- Development of a fusion method for synchronizing LiDAR and thermal imaging, overcoming modality differences to enable accurate alignment of thermal anomalies onto 3D models.

- Development of a real-time thermal imaging and cloud points reconstruction model through a novel Thermal-Lio system that captures thermal information from multiple viewpoints, overcoming the limitations of conventional outdoor thermography SHM tasks.

- Development of an approach that can effectively demonstrate detailed temperature variations and temporal evolution in outdoor structures, thereby providing valuable insights for future building design, maintenance planning, and structural health management.

The proposed Thermal-Lio system successfully overcomes the inherent limitations of traditional two-dimensional thermography, enhancing defect detection precision, providing extensive spatial coverage, and enabling continuous temperature monitoring. Future research may focus on optimizing real-time data processing algorithms.